\documentclass[a4paper]{article}

\usepackage{INTERSPEECH2021}

\usepackage{array}
\newcolumntype{L}[1]{>{\raggedright\arraybackslash}p{#1}}
\newcolumntype{C}[1]{>{\centering\arraybackslash}p{#1}}
\newlength\cwidth
\usepackage[table,x11names]{xcolor}
\usepackage[hyphens,spaces,obeyspaces]{url}
\title{Bridging the gap between streaming and non-streaming ASR systems by distilling ensembles of CTC and RNN-T models}
\name{Thibault Doutre, Wei Han, Chung-Cheng Chiu, Ruoming Pang, Olivier Siohan, Liangliang Cao}
\address{Google Inc., USA}
\email{doutre@google.com, llcao@google.com}

\begin{document}

\maketitle

\begin{abstract}
Streaming end-to-end automatic speech recognition (ASR) systems are widely used in everyday applications that require transcribing speech to text in real-time. Their minimal latency makes them suitable for such tasks. Unlike their non-streaming counterparts, streaming models are constrained to be causal with no future context and suffer from higher word error rates (WER). 
To improve streaming models, a recent study \cite{baseline} proposed to distill a non-streaming teacher model on unsupervised utterances, and then train a streaming student using the teachers' predictions. 
However, the performance gap between teacher and student WERs remains high. In this paper, we aim to close this gap by using a diversified set of non-streaming teacher models and combining them using Recognizer Output Voting Error Reduction (ROVER). In particular, we show that, despite being weaker than RNN-T models, CTC models are remarkable teachers. Further, by fusing RNN-T and CTC models together, we build the strongest teachers. The resulting student models drastically improve upon streaming models of previous work \cite{baseline}: the WER decreases by 41\% on Spanish, 27\% on Portuguese, and 13\% on French. 
\end{abstract}

\section{Introduction}

ASR systems are used to transcribe speech to text in many daily applications. They can be embedded on devices like smart home devices or smartphones and can also be used in cloud services. These systems can be designed to be either steaming or non-streaming. Non-streaming models can take advantage of the full sequence of audio when transcribing speech \cite{attentionbased,las}. However, this requires that the full speech sequence be sent to the system before having access to a transcript. Streaming end-to-end ASR systems \cite{graves,he2019streaming,zhang2020transformer,yeh2019transformer,tsunoo2019towards,moritz2020streaming,li2020towards} are developed for real-time recognition tasks such as smart assistants and real-time captioning. Due to their streaming constraint, such systems cannot use the full context and generally perform less than non-streaming systems. This paper addresses this performance gap between streaming and non-streaming models.

Recent research has focused on improving streaming ASR by trading off latency \cite{sainath2019two,las,sainath2020streaming}. For example, \cite{lookahead} shows that allowing streaming models to add a small amount of future information is effective in reducing the WER of RNN-T models. Some work attempts on improving streaming ASR using model distillation \cite{ctcdistillation1,ctcdistillation2,ctcdistillation3,Saon-distillation-2020}. More recently, \cite{baseline} extends this approach by distilling a non-streaming teacher model on large-scale unsupervised YouTube dataset. By scaling the unsupervised data set, streaming models are trained on a large amount of training data and can outperform streaming models trained on a limited amount of supervised data. However, there is still a significant gap between teacher and student WERs.

\begin{figure}[t]
\includegraphics[width=6.5cm]{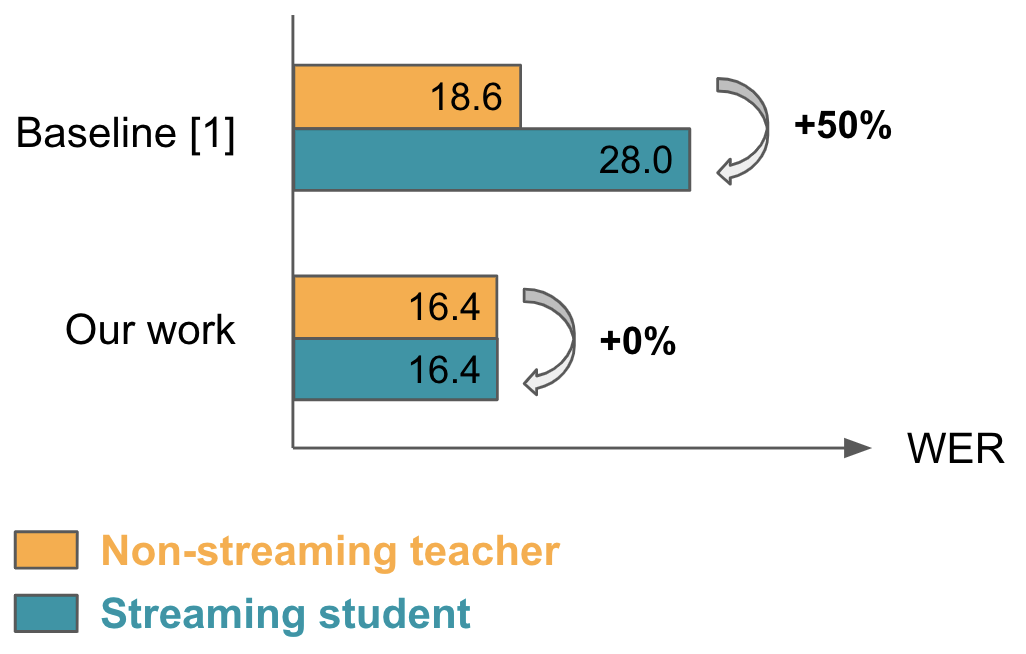}
\caption{Comparing WER between streaming students and their respective teachers (Spanish). The gap in performance between teachers and students WERs significantly improves over previous work \cite{baseline}.} 
\label{fig:gap}
\end{figure}
\begin{figure}[t]
\includegraphics[width=7.5cm]{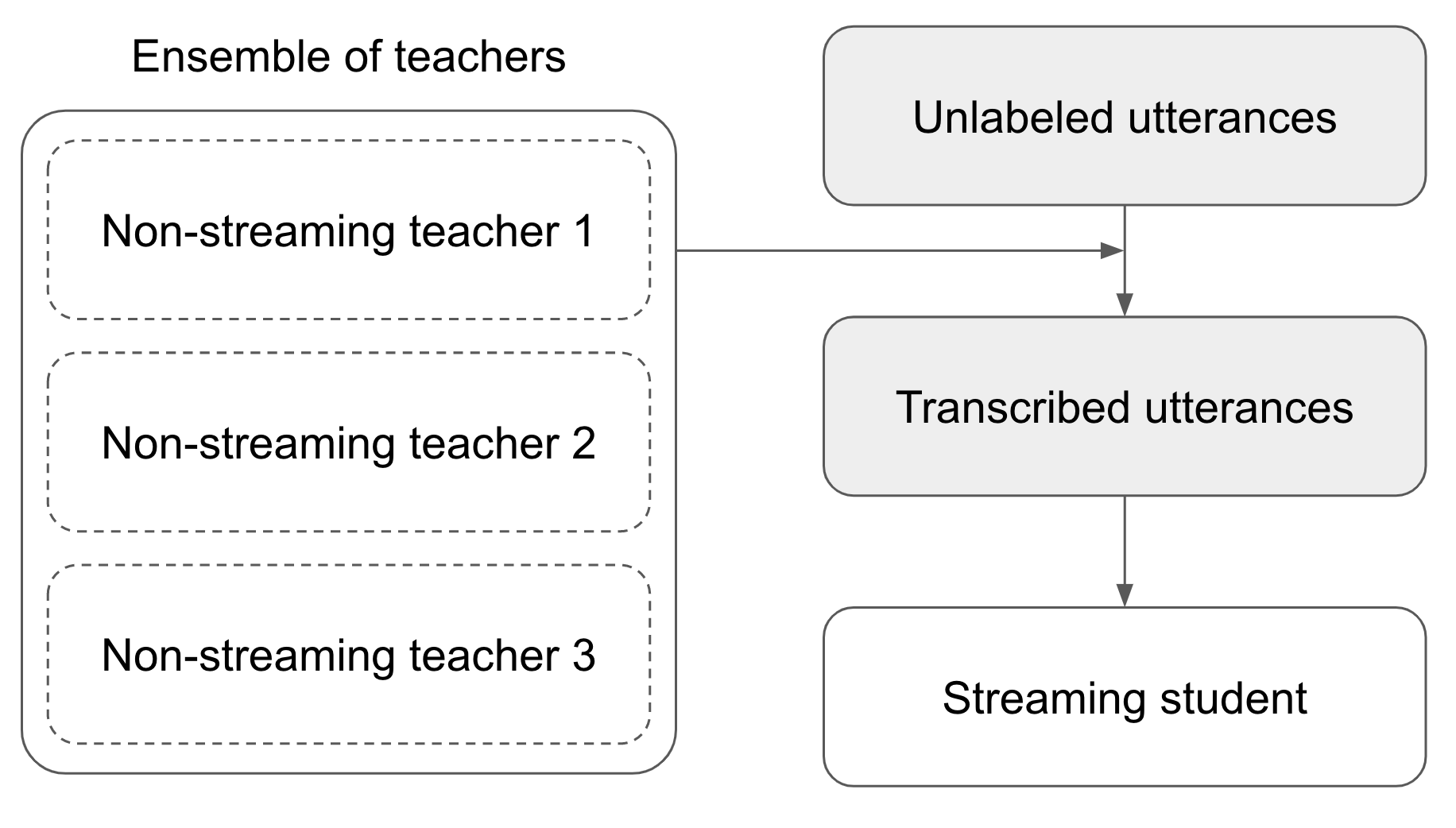}
\caption{Teacher-student framework. The student model is trained on an arbitrarily large set of utterances, transcribed by an ensemble of pre-trained teacher models.}
\label{fig:tsl}
\end{figure}

In this paper, we aim to further reduce the gap between non-streaming teachers and streaming students. We propose to improve this approach by (1) using a variety of non-streaming teacher models (2) combining teacher models to improve the quality of transcription (3) using stronger student models. In particular, we find Connectionist Temporal Classification (CTC) models \cite{ctc}  better teacher models than RNN-T models. In previous literature, RNN-T was considered better than CTC \cite{he2019streaming}, as CTC suffers from the frame-independence assumption and often relies on an additional language model to capture the context information. In this work, however, RNN-T student models appear to generalize better when trained from CTC teachers. In addition, we find it advantageous to combine CTC with RNN-T, as the ensemble brings significant gains: the diversity of such models creates stronger teachers than any single model. 

When comparing our work to \cite{baseline}, we drastically improve the performance of streaming models using a similar amount of labeled data in multiple languages (later shown in Table \ref{tab:baseline_wer}). According to \cite{baseline}, the WER increase of student models ranges from 24\% to 50\% relative to their respective teachers, depending on the language. Figure \ref{fig:gap} shows that the performance gap between streaming students and non-streaming teachers is closed in Spanish. In section \ref{sec:experiments}, we will show that this behavior is consistent in all languages.  

\section{Method}

\subsection{The teacher-student framework}
In \cite{baseline}, the authors describe a method to improve streaming ASR by using non-streaming models. Given a pre-trained non-streaming teacher model, a set of randomly segmented YouTube audio data is transcribed. The resulting data set is then augmented with pseudo labels and used to train streaming models. The authors showed that the teacher's performance is an important factor in the student's performance. By scaling this method to millions of hours of unlabeled data, streaming models improve up to 39\% compared to a baseline RNN-T model.

\subsection{Ensemble of teachers}
\begin{figure}[t]
\includegraphics[width=8cm]{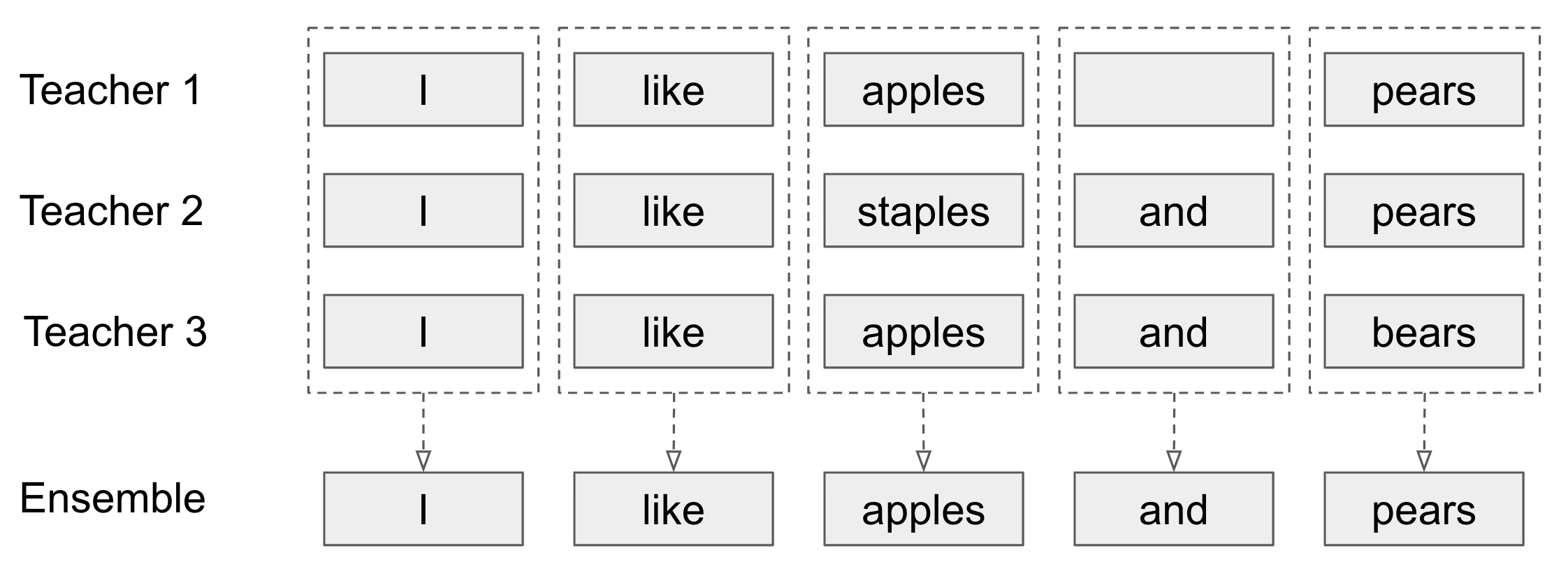}
\caption{Transcripts of multiple teacher models combined in a transition network by aligning them iteratively. The resulting transcript is obtained with majority voting.}
\label{fig:wtn}
\end{figure}

To improve the teacher-student framework, we propose to train several teacher models and combine them with ROVER \cite{rover}. The data is first transcribed with multiple teachers. The resulting transcripts are iteratively aligned to form a word transition network (WTN). Then the edges of the WTN are merged by majority voting, as described in Figure \ref{fig:wtn}. Since the resulting transcripts come from the predictions of several teachers, they can be viewed as the transcription of an ensemble of teacher models. Should the ensemble of teachers perform better than any single teacher, the resulting transcripts may also be better to learn from. 

\subsection{Combining RNN-T and CTC models}
ROVER is known to perform best when using various models, so it can exploit differences in the categories of errors generated by the different ASR systems \cite{rover2009, rover2000}. In this paper, we choose to combine multiple RNN-T \cite{graves} and CTC \cite{ctc} models. For all teacher models, we adopt a Conformer \cite{conformer} based encoder which has shown improvements over alternatives by taking advantage of both convolution neural networks (CNN) and transformers’ architectures to capture local and global context. 

In all languages, we consider three teacher models. The first non-streaming teacher is the \textbf{MD-RNNT} model \cite{conformer}. This model stacks 17 macro layers in the encoder. Each macro layer consists of 8 attention heads, 1-D convolutions with a kernel size of 15 and a relative positional embedding of dimension 512. We use a right context of size 42 and a left context of size 43. The encoder dimension is 512. The decoder network has a uni-directional LSTM with 1024 hidden units. The joint network has 512 hidden units and the final output uses a 4k word piece model. The total number of parameters for this model is 141 million. The second teacher model we are considering is the \textbf{YT-RNNT} model \cite{tdnn}. The encoder starts with 2 layers of 3x3 2D convolution layers at the bottom with 4x time reduction, with a channel size of 640. There are a total of 16 conformer blocks, that are similar to the macro layers of MD-RNNT, except for the convolutions which have a kernel size of 32 and a positional embedding of size 128. The encoder dimension is 640. The decoder layer has 1 LSTM with 2048 cells and a projection layer of 640 outputs. This makes up a total of 175 million parameters. For both MD-RNNT and YT-RNNT models, we follow \cite{conformer} to set the front-end and SpecAugment hyper-parameters. The third teacher model used in this study is a CTC model: \textbf{YT-CTC}. It has the same encoder as the YT-RNNT model, but its decoder only has 1 layer of 4096 hidden units with a projection layer of 640 outputs. Due to the simplified nature of the encoder, this model has 167 million parameters which is fewer than the YT-RNNT model. Both YT-CTC and YT-RNNT models are trained from YouTube data, whereas MD-RNNT is trained on multi-domain data \cite{mutlidomain}. The data include audio from YouTube, but also other data sources that come from user requests made on various devices.

As for our student model, we use the same streaming RNN-T model with the Conformer encoder across all languages. This model is similar to the MD-RNNT, with the exception that attention layers and convolutions no longer allow full context to ensure its streaming quality.

\begin{table}[t]
\caption{Architecture of the different teacher models.}
\begin{tabular}{ |p{0.3\cwidth}|C{0.2\cwidth}|C{0.2\cwidth}|C{0.3\cwidth}| } 
\hline
 & Encoder & Decoder & Data\\ 
\hline
MD-RNNT & 17 blocks & 1 LSTM & Multi-domain\\ 
YT-RNNT & 16 blocks & 1 LSTM & YouTube \\ 
YT-CTC & 16 blocks & 1 layer & YouTube \\ 
\hline
\end{tabular}
\label{tab:teacher_models}
\end{table}

\section{Experiments on YouTube data} \label{sec:experiments}

We confirm our method on YouTube data in three different languages: Spanish, Portuguese, and French. For each language, we are given three non-streaming teacher models, later combined with ROVER. To ensure a fair comparison, we use the same streaming RNN-T student model in all experiments.

\subsection{Evaluation set}
We evaluate our models on \textit{YT-long}, a set of utterances generated by sampling and hand-transcribing popular YouTube videos based on their view counts. The length of utterances varies between 40 seconds and 30 minutes. When decoding an utterance, no segmentation is applied: the whole utterance is being processed at once.

\subsection{Training data}
As in \cite{baseline}, we train our teacher models on \textit{Confisland} data \cite{confisland}. The data is semi-supervised and has been acquired from a set of videos with transcripts uploaded by YouTube video owners. Then, as described in \cite{confisland}, only part of the audio where the transcript matches the audio with a certain confidence is being kept for training. Approximately 30\% to 50\% of the audio is kept, depending on the language. 

As for our unsupervised set \textit{YT-segments}, we randomly segment the audio of the original set of \textit{Confisland} into utterances of lengths varying uniformly between 5 and 15 seconds. Since \textit{YT-segments} are unlabeled, we can use all the data, even if the user uploaded transcripts are missing or incorrect. The size of each data set is described in Table \ref{tab:data_training}.

\begin{table}[t]
\caption{Number of hours of the \textit{Confisland} and \textit{YT-segments} data sets for different languages.}
\begin{tabular}{ |p{0.35\cwidth}|C{0.35\cwidth}|C{0.35\cwidth}| } 
\hline
 & \textit{Confisland} & \textit{YT-segments} \\ 
\hline
Spanish & 13,000  & 41,000   \\ 
French& 10,000  & 29,000 \\ 
Portugese & 2,500 & 5,000   \\ 
\hline
\end{tabular}
\label{tab:data_training}
\end{table}

\subsection{Results}
\begin{table}[t]
\centering
\caption{WERs of a streaming Conformer student model trained on \textit{YT-segments}, distilled from non-streaming teacher models.}
\begin{tabular}{ |p{0.2\cwidth}|p{0.4\cwidth}|C{0.2\cwidth}|C{0.2\cwidth}| }
    \hline
     & Teacher model & Teacher WER on \textit{YT-long}& Student WER on \textit{YT-long} \\
    \hline
    Spanish & MD-RNNT &  16.4 &33.4 \\
     & YT-RNNT &  18.6 &23.4  \\
     & YT-CTC &  20.2 &16.9  \\
     & Teacher ensemble &  18.1 &16.4  \\
    \hline
    Portuguese & MD-RNNT &  29.1 &31.9 \\
     & YT-RNNT &  22.8 &26.7  \\
     & YT-CTC &  24.8 &23.0  \\
     & Teacher ensemble &  21.9 &20.5  \\
    \hline
    French & MD-RNNT &  31.9 & 42.8\\
     & YT-RNNT &  18.8 &23.6  \\
     & YT-CTC &  21.0 &16.6  \\
     & Teacher ensemble &  20.2 &16.7  \\
    \hline
\end{tabular}
\label{tab:student_wer}
\end{table} 
\begin{table}[t]
\caption{Relative change of streaming students WERs, compared to their respective non-streaming teachers. The lower the increase the better. When multiple teachers are involved, the WER of the best teacher is used.}
\begin{tabular}{ |p{0.18\cwidth}|C{0.43\cwidth}|C{0.45\cwidth}| } 
\hline
 & Baseline \cite{baseline} & Our streaming student \\ 
\hline
Spanish & +50\% & +0\%  \\ 
Portugese & +24\% & -6\%   \\ 
French& +34\% & -11\% \\ 
\hline
\end{tabular}
\label{tab:teacher_student_increase}
\end{table}

In Table \ref{tab:student_wer} we compare how well the same streaming RNN-T model performs when trained from different teachers and from the ensemble of teachers. We note that, despite the higher WER of CTC models, they are better teachers than their non-streaming RNN-T counterparts. In particular, RNN-T students trained from CTC teachers surpass their teachers on \textit{YT-long}. When using the model ensemble, teacher performance not only improves, but also its corresponding student. The gap between the best streaming student and the best non-streaming teacher is significantly reduced.

Specifically, Table \ref{tab:teacher_student_increase} and Table \ref{tab:baseline_wer} show the improvement over the results in \cite{baseline}: by using an ensemble of teachers and a RNN-T student model, we outperform baselines trained from single non-streaming RNN-T teachers on similar data. The relative improvement in WER varies between 27\% for Portuguese and 42\% for Spanish.


\begin{table}[t]
\caption{Comparison of the WER of streaming models in this paper compared with streaming baselines \cite{baseline} trained on similar data.}
\begin{tabular}{ |p{0.45\cwidth}|c|c|c| } 
\hline
 & Spanish & Portuguese & French\\ 
\hline
Streaming RNN-T on \textit{Confisland} \cite{baseline} & 35.9 & 30.8  & 34.5 \\ 
\hline
Baseline streaming student \cite{baseline} & 28.0 & 28.3 & 25.0 \\ 
\hline
Our streaming student & 16.4 & 20.5 & 16.7 \\ 
\hline
Relative improvement relative to the baseline streaming student & 41\% & 27\% & 13\% \\
\hline
\end{tabular}
\label{tab:baseline_wer}
\end{table}

\subsection{CTC vs RNN-T teachers}

\begin{figure}[!b]
\includegraphics[width=7.5cm]{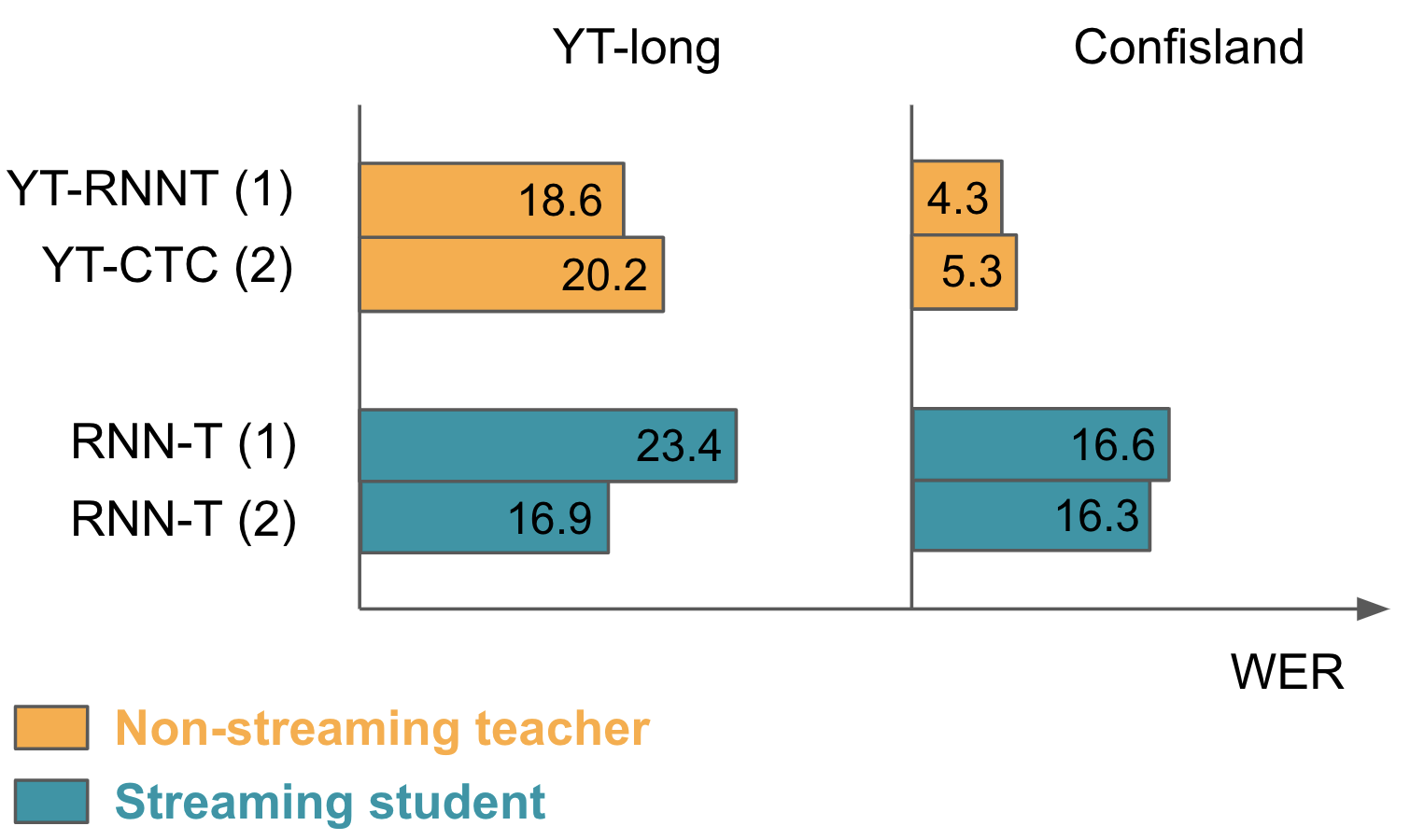}
\caption{Comparing a RNN-T streaming student model trained from CTC and RNN-T non-streaming teachers. The CTC model is a better teacher, despite having a higher WER.} 
\label{fig:ctc_rnnt}
\end{figure}

Figure \ref{fig:ctc_rnnt} and Table \ref{tab:student_wer} show that student models trained from CTC and RNN-T teachers behave differently. When CTC is used, the corresponding student models outperform their teachers on \textit{YT-long}. However, when RNN-T teachers are used, students preform worse than their teachers. Looking at several teacher predictions on the French \textit{YT-segments}, we found that CTC transcripts are generally worse than their RNN-T counterparts. Table \ref{tab:audit} shows a typical utterance from \textit{YT-segments}, transcribed using multiple teachers. The CTC transcript suffers from grammatical problems and made-up words. This is likely a result of the local nature of the CTC decoder, not being able to form a coherent sentence from word pieces.

Despite generating worse labels, the CTC model appears to be a better teacher to learn from. To determine whether this behavior is tied to \textit{YT-long}, we look at the WER on a \textit{Confisland} dev set: across all languages and for all models, students always perform worse than their teachers. Table \ref{tab:confisland_wer} shows such gap between teachers and students when evaluated on \textit{Confisland} data. The main difference between \textit{Confisland} and \textit{YT-long} being the length of utterances, it seems that student models trained from CTC teachers suffer less from long form errors. 

\begin{figure}[t]
\includegraphics[width=8cm]{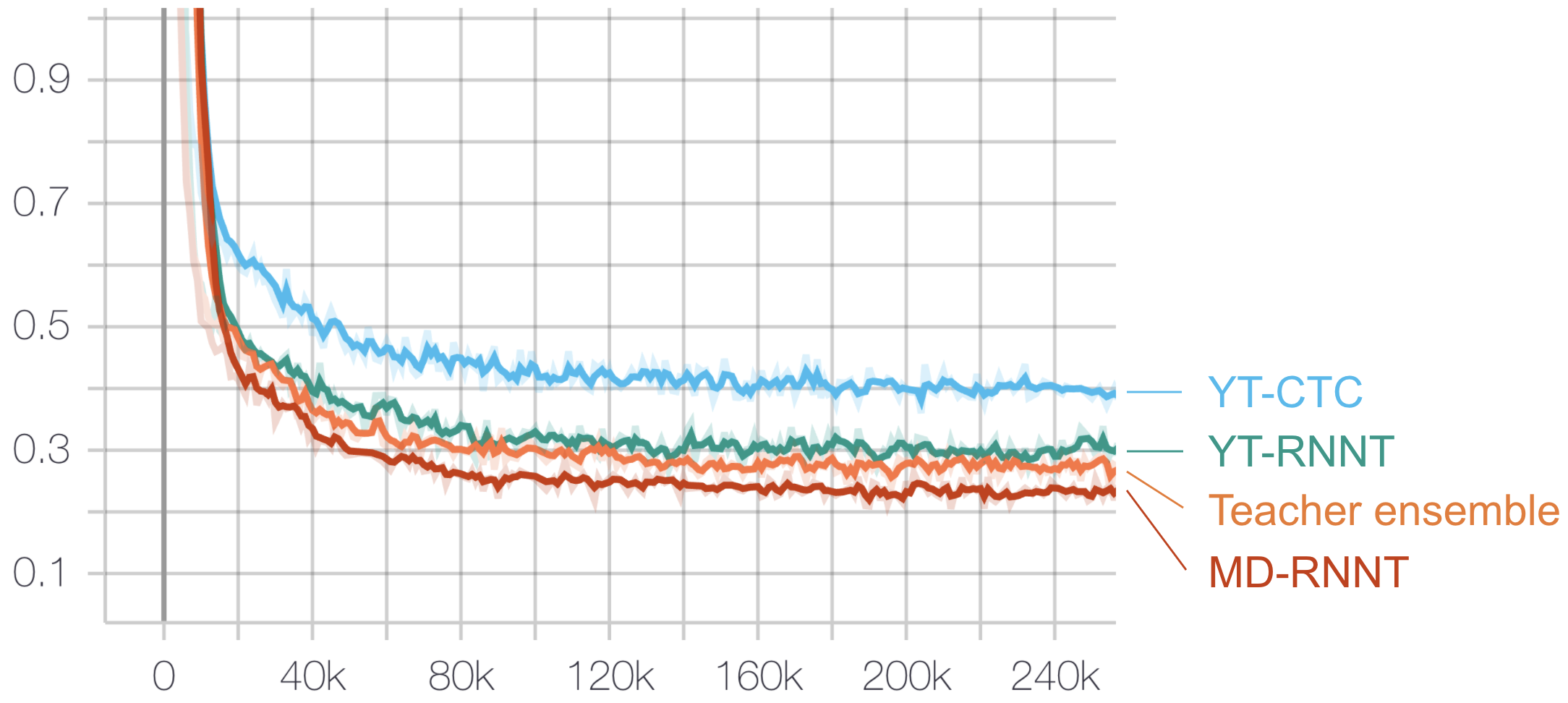}
\caption{Loss of student models trained from different teachers on the Spanish \textit{Confisland} dev set, normalized per token.} 
\label{fig:loss}
\end{figure}

Finally, we look at the students' training losses, normalized per token. Figure \ref{fig:loss} shows that students trained from CTC teachers have a higher loss than those trained from RNN-T teachers. This indicates that RNN-T transcripts are easier to fit than CTC transcripts. The high loss of students trained from CTC teachers and their low WERs on \textit{YT-long} suggests that students trained from CTC teachers generalize better on long utterances. The student-teacher framework could be more robust when teachers and students are of different nature. Such behavior has already been observed in the experiments of~\cite{yu2020}, where the authors compare two different student models for the same ContextNet teacher \cite{contextnet} in a noisy student setting \cite{noisy_student}. The Conformer student has better results than the ContextNet student on Librispeech \cite{librispeech}.

\begin{table}[t]
\caption{Comparing different teachers' predictions on French \textit{YT-long}. Bold words are not in the French vocabulary and underlined words are not in the ground truth.}
\begin{tabular}{ |p{0.24\cwidth}|p{0.9\cwidth}| } 
\hline
Teacher model & Prediction \\ 
\hline
MD-RNNT & \underline{euh} qui \underline{achète} pour revendre sur les vide-greniers sans s'acquitter des taxes et obligations     \\ 
\hline
YT-RNNT& \underline{gens} qui achètent pour revendre sur les vide-greniers sans s'acquitter des taxes et obligations \underline{ou}   \\ 
\hline
YT-CTC & qui \underline{\textbf{achèèteent}} pour revendre sur \underline{le} \underline{\textbf{videe-} grenier} sans accquitter des taxes et obligations \underline{ou}   \\ 
\hline
Teacher ensemble & \underline{gens} qui pour revendre sur \underline{le} vide-greniers sans s'acquitter des taxes et obligations ou   \\ 
\hline
Ground truth & qui achètent pour revendre sur les vide-greniers sans s'acquitter des taxes et obligations   \\ 
\hline
\end{tabular}
\label{tab:audit}
\end{table}
\begin{table}[t]
\centering
\caption{WERs of a streaming Conformer student model and their teachers on a Spanish \textit{Confisland} dev set.}
\begin{tabular}{ |p{0.38\cwidth}|C{0.35\cwidth}|C{0.35\cwidth}| }
    \hline
    Teacher model & Teacher WER on \textit{Confisland}& Student WER on \textit{Confisland} \\
    \hline
    MD-RNNT &  6.8 &17.8 \\
     YT-RNNT &  4.3 &16.6  \\
     YT-CTC &  5.3 &16.3  \\
     Teacher ensemble &  4.1 &16.2  \\
    \hline
\end{tabular}
\label{tab:confisland_wer}
\end{table}

\subsection{Ablation studies}

In this subsection, we aim to better understand how RNN-T and CTC teachers work differently with student models.

\begin{table}[t]
\caption{WERs of teachers and their respective students in different ablation studies for the Spanish \textit{YT-long} test set.}
\begin{tabular}{ |p{0.38\cwidth}|p{0.32\cwidth}|C{0.15\cwidth}|C{0.15\cwidth}| } 
\hline
Teacher ensemble & RNN-T Student & Teacher WER & Student WER \\ 
\hline
RNN-T + CTC & Streaming &18.1  & 16.4   \\ 
RNN-T + CTC & Non-streaming & 18.1& 14.7   \\ 
RNN-T only & Streaming &17.5 & 20.1   \\ 
CTC only & Streaming &19.6 & 18.0   \\ 
\hline
\end{tabular}
\label{tab:ablations}
\end{table}


\subsubsection{A non-streaming student}
First, we train a non-streaming version of the student model on the Spanish \textit{YT-segments} distilled from the ensemble of teachers. This non-streaming student has the same architecture as MD-RNNT and improves the WER by 10\% compared to its streaming counterpart (Table \ref{tab:ablations}). This underlines that distillation itself contributes to improving non-streaming RNN-T models.

\subsubsection{An ensemble of RNN-T teachers}
In this subsection, we examine the importance of CTC in the ensemble of teachers. Specifically, we replace the YT-CTC teacher with another version of MD-RNNT. The WER of the streaming student trained from this new ensemble of RNN-T models increases by 19\% compared to when using the diverse set of teachers (Table \ref{tab:ablations}). This highlights the fact that at least one CTC model as a teacher is critical to the student's performance.

\subsubsection{An ensemble of CTC teachers}
Lastly, we provide evidence that a diverse set of teachers is beneficial for student models. We train student models from an ensemble of CTC teachers made from five different checkpoints of YT-CTC. The performance of such teachers ranges from 20.2 to 21.0. The resulting ensemble improved the WER to 19.6. Table \ref{tab:ablations} shows, however, that the student model that is trained from this resulting ensemble of CTC teachers does not perform as well as with the diverse group of teachers: the WER increases from 16.4 to 18.0.

\section{Conclusions}

In this paper, we develop a robust teacher-student learning method.  In particular, we observe that CTC teachers, despite having a higher WER than their RNN-T counterpart, are a better teacher for streaming RNN-T models. 
By combining CTC and RNN-T models into a teacher ensemble, we can train a new streaming student model that not only drastically improves the results in \cite{baseline}, but also close the gap between single non-streaming teachers and streaming students. Future work will include improving the CTC teacher by complementing it with a language model, expanding the amount of unsupervised data and adding more teachers to the ensemble.
\section{Acknowledgements}

We are very thankful for our colleagues Basi García, Min Ma, Zhiyun Lu, Yongqiang Wang, Anmol Gulati, Françoise Beaufays, and Trevor Strohman for their help and suggestions.
 
\bibliographystyle{IEEEtran}
\bibliography{mybib}

\end{document}